\documentclass[runningheads]{llncs}
\usepackage{graphicx}

\usepackage{cite}
\usepackage{tikz}
\usepackage{comment}
\usepackage{amsmath,amssymb} 
\usepackage{color}

\usepackage{hyperref}
\usepackage{url}
\usepackage[inline]{enumitem}

\usepackage[utf8]{inputenc} 
\usepackage[T1]{fontenc}    
\usepackage{hyperref}       
\usepackage{url}            
\usepackage{booktabs}       
\usepackage{amsfonts}       
\usepackage{nicefrac}       
\usepackage{microtype}      
\usepackage{xcolor}         
\usepackage{graphicx}
\usepackage{xcolor,colortbl}
\usepackage{multirow}
\usepackage{caption}
\usepackage{wrapfig}
\usepackage{subcaption}
\definecolor{srcolor}{rgb}{0,0,0}
\newcommand{\sr}[1]{\textcolor{srcolor}{{#1}}}
\DeclareMathOperator{\loss}{\mathcal{L}}
\DeclareMathOperator{\E}{\mathbb{E}}

\usepackage{xcolor,colortbl}

\usepackage{mathtools}
\definecolor{Gray}{gray}{0.85}
\definecolor{LightCyan2}{rgb}{0.94,0.81,0.81}

\definecolor{LightCyan}{rgb}{0.84,0.84,0.84}

\def\E{{\rm E}\,}

\usepackage[accsupp]{axessibility}  

\begin{document}
\pagestyle{headings}
\mainmatter
\def\ECCVSubNumber{12}  

\title{Unsupervised Scene Sketch to Photo Synthesis} 

\titlerunning{Unsupervised Scene Sketch to Photo Synthesis}
%
\author{
\setlength{\tabcolsep}{10mm}
\begin{tabular}{@{}ccc@{}}
Jiayun Wang\inst{1} &
Sangryul Jeon\inst{1} &   
Stella X. Yu\inst{1} \\
Xi Zhang\inst{2} &
Himanshu Arora\inst{2} &
Yu Lou\inst{2} \\
\end{tabular}}
\authorrunning{Wang et al.}
\institute{$^1$ UC Berkeley / ICSI \hspace{10em}  $^2$ 
Amazon\\
\email{\{peterwg,srjeon,stellayu\}@berkeley.edu \hspace{3mm} \{xizhn,arorah,ylou\}@amazon.com}
}

\maketitle
\begin{abstract}

Sketches make an intuitive and powerful visual expression as they are fast executed freehand drawings. We present a method for synthesizing realistic photos from scene sketches. Without the need for sketch and photo pairs, our framework directly learns from readily available large-scale photo datasets in an unsupervised manner. 
To this end, we introduce a standardization module that provides {\it pseudo} sketch-photo pairs during training by converting photos and sketches to a standardized domain, i.e. the edge map.
\sr{The reduced domain gap between sketch and photo also allows us to disentangle them into two components: holistic scene structures and low-level visual styles such as color and texture.}
\sr{Taking this advantage, we synthesize a photo-realistic image by combining the structure of a sketch and the visual style of a reference photo.}
Extensive experimental results on perceptual similarity metrics and human perceptual studies show the proposed method could generate realistic photos with high fidelity from scene sketches and outperform state-of-the-art photo synthesis baselines.
We also demonstrate that our framework facilitates a controllable manipulation of photo synthesis by editing strokes of corresponding sketches, delivering more fine-grained details than previous approaches that rely on region-level editing.

\keywords{sketch, scene sketch, photo synthesis, unsupervised learning}

\end{abstract}
\begin{figure}[t!]\centering
\centering
\vspace{-1em}
  \includegraphics[width=\textwidth]{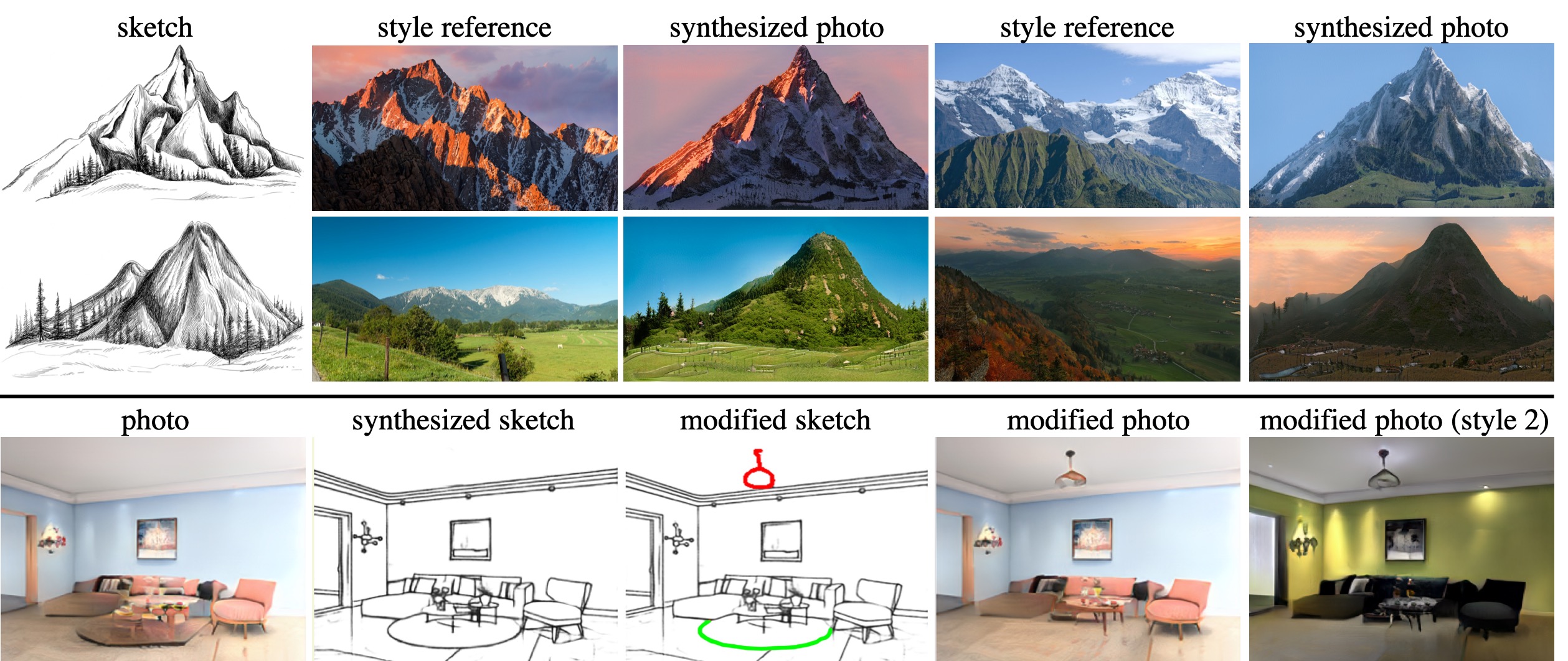}
  \vspace{-1em}
  \caption{
  {\it Upper:} Given a sketch and a style reference photo, our method is capable of transferring low-level visual styles of the reference while preserving the content structure of the sketch. We show synthesis results with different references. {\it Lower:} Given an arbitrary photo, users could easily and interactively edit it by  \textcolor{red}{adding} or \textcolor{green}{removing} strokes on the synthesized sketch.
  }
  \vspace{-1em}
  \label{fig:teaser}
\end{figure}

\vspace{-1em}
\section{Introduction}

Sketching is an intuitive way to represent visual signals. With a few sparse strokes, humans could understand and envision a photo from a sketch. Additionally, unlike photos which are rich in color and texture, sketches are easily editable as strokes are easy to modify.
\sr{We aim to synthesize photos that preserve the structure of scene sketches while delivering the low-level visual style of reference photos.}



\sr{Unlike previous works~\cite{isola2017image,liu2020unsupervised,richardson2021encoding} that synthesize photos from categorical object-level sketches, 
our goal in which scene-level sketches are used as input poses additional challenges due to} 
{\bf 1) Lack of data.} There is no training data available for our task due to the complexity of scene sketches. Not only the insufficient amount of scene sketches, but the lack of paired scene sketch-image datasets make supervised learning from one modality to another intractable.
{\bf 2) Complexity of scene sketches.} A scene sketch usually contains many objects of diverse semantic categories with complicated spatial organization and occlusions. Isolating objects, synthesizing object photos and combining them together \cite{gao2020sketchycoco} do not work well and are hard to generalize. For one, detecting objects from sketches is hard due to the sparse structure. For another, one may encounter objects that do not belong to seen categories, and the composition could also make the synthesized photo unrealistic.



We propose to alleviate these issues via {\bf 1)} a standardization module, and {\bf 2)} disentangled representation learning.

For the lack of data,
we propose a standardization module, where input images are converted to a standardized domain, edge maps. Edge maps can be considered as {\it synthetic sketches} due to the high similarity to real sketches. With the standardization, readily-available large-scale photo datasets could be used for training by converting them to edge maps. Additionally, during inference, sketches of various individual styles are also standardized such that the gap between training and inference is narrowed.

For the complexity of scene sketches, we learn disentangled holistic content and low-level style representations from photos and sketches by encouraging only content representations of photo-sketch pairs to be similar.
As a definition, content representations encode holistic semantic and geometric structures of a sketch or photo. Style representations encode the low-level visual information such as color and texture.
A sketch could depict similar contents as a photo, but contain no color or texture information. 
By factorizing out colors and textures, the model could directly learn from large-scale photos for scene structures and transfer the knowledge to sketches.
Additionally, combining the content representation of a sketch and a style representation of a reference photo could decode a realistic photo. The decoded photo should depict similar contents as the sketch and shares a similar style with the reference photo.  This is the underlying mechanics of the proposed reference-guided scene sketch to photo synthesis approach. Note that the disentangled representations have been studied previously for photos \cite{viazovetskyi2020stylegan2,park2020swapping} and we extend the concept to sketches.

As exemplified in Fig.\ref{fig:teaser}, not only photo synthesis from scene sketch, our model can promote also controllable photo editing by allowing users to directly modify strokes of a corresponding sketch. The process is easy and fast as strokes are easy and flexible to modify, compared with photo editing from segmentation maps proposed by previous works \cite{ling2021editgan,isola2017image,meng2021sdedit,park2020swapping}. Specifically, the standardization module first converts a photo to a sketch. Users could modify strokes of the sketch and synthesize a newly edited photo with our model. Additionally, the style of the photo could also be modified with another reference photo as guidance.


We summarize our contribution as follows:
		{\bf 1)} We propose an unsupervised scene sketch to photo synthesis framework. We introduce a standardization module that converts arbitrary photos to standardized edge maps, enabling a vast amount of real photos to be utilized during training.
        {\bf 2)} Our framework facilitates controllable manipulation of photo synthesis through editing scene sketches with more plausibility and simplicity than previous approaches.
        {\bf 3)} Technically, we propose novel designs for scene sketch to photo synthesis, including shared content representations to enable knowledge transfer from photos to sketches and model fine-tuning with sketch-reference-photo triplets for improved performance. 
\vspace{-10pt}

\begin{figure}[t!]\centering
\vspace{-3pt}
\includegraphics[width=\columnwidth]{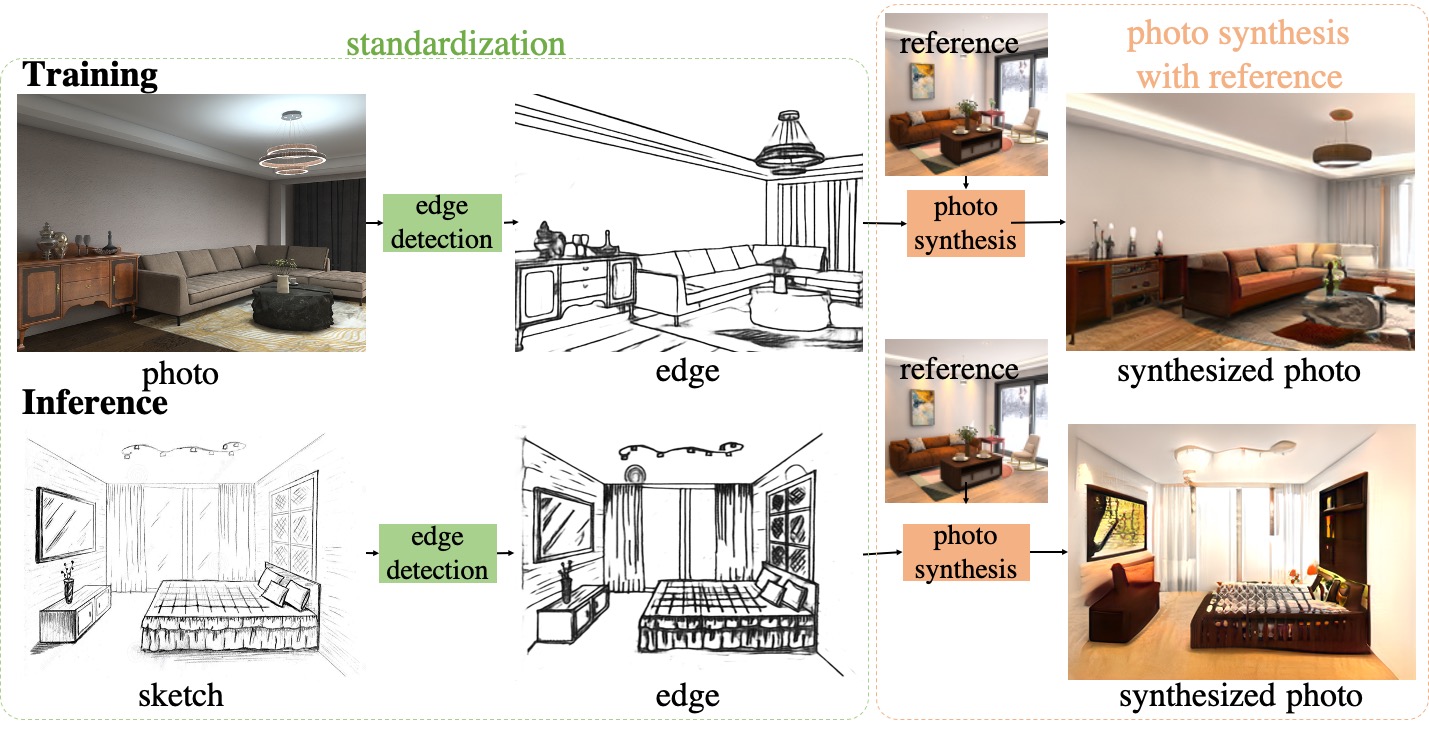}
\vspace{-1em}
\caption{ 
Our method consists of two components, standardization and photo synthesis. {\bf Left: } The standardization module converts photos or sketches into a standardized domain, edge maps,  to reduce the domain gap between training and inference. {\bf Right: } From the standardized edge map, the photo synthesis module generates a photo with a similar style as the given reference image. }
\vspace{-10pt}
\label{fig:overview}
\end{figure}
\section{Related Work}

\sr{\textbf{Conditional Generative Models}.}
Previous approaches generated realistic images by conditioning generative adversarial networks~\cite{goodfellow2014generative} on a given input from users. More recent methods extended it to multi-domain and multi-modal setting~\cite{liu2019few,huang2018multimodal,choi2020stargan}, facilitating numerous downstream applications including image inpainting~\cite{iizuka2017globally,pathak2016context}, photo colorization~\cite{zhang2016colorful,larsson2016learning}, texture and geometry synthesis~\cite{zhou2018non,guerin2017interactive}.
However, naively adopting this framework to our problem is challenging due to the absence of paired data where sketches and photos aligned.
We address this by projecting arbitrary sketches and photo into the intermediate representation and generating pseudo paired data to learn in an unsupervised setting.

\noindent \textbf{Disentanglement of Content and Style Representations.} The disentanglement has been studied \cite{zhu1998filters,portilla2000parametric} prior to the surge of deep learning models, where they show low-level style like texture can be modeled as statistics of an image.
Deep generative models \cite{viazovetskyi2020stylegan2,karras2019style,park2020swapping,lee2020drit++} also achieved success in photo style transfer by the disentanglement.
We extend the disentanglement idea to sketches and show its application in photo synthesis.

\noindent \sr{\textbf{Sketch to Photo Synthesis}.}
Following a seminal work, SketchGAN~\cite{chen2018sketchygan}, several efforts has been made on synthesizing photos\cite{ghosh2019interactive,liu2020unsupervised,xiang2022adversarial} or reconstructing 3D shapes~\cite{wang20203d,delanoy20183d,wang2018unsupervised} from sketches.
They however mainly focused on categorical single-object sketches without substantial background clutters, and thus have difficulties when encountered with complicated scene-level sketches. 

Scene sketch to photo synthesis is limited by lack of the data. SketchyScene \cite{zou2018sketchyscene} is the only scene dataset with object segmentation and corresponding cartoon images. However, their sketch is manually composited from multiple object sketches with reference to a cartoon image. The composite sketch has a large domain gap to real scene sketches with reference to a real scene. Their composition idea greatly impacts how researchers solve the photo synthesis. \cite{gao2020sketchycoco} detect objects of composite sketches and generate individual photos as well as a background image and combine them together. Holistic scene structures are ignored and the photo composition leads to artifacts and unrealism. We learn holistic scene structures from massive photo datasets and transfer the knowledge to sketches.

\noindent \sr{\textbf{Deep Image Editing}.}
By the favor of powerful generative models~\cite{karras2020analyzing}, previous works edited photos by modifying the extracted latent vector. Typically they sampled the desired latent vector from a fixed distribution according to a user's semantic control~\cite{zhu2016generative}, or let a user spatially annotate the region-based semantic layout~\cite{park2020swapping,park2019semantic}.
DeepFaceDrawing \cite{chen2020deepfacedrawing} enables user to  sketch progressive for face image synthesis.
Our work differs in that we allow users to directly edit strokes of a complicated scene sketch, thus enabling much more fine-grained editing.

\section{Methods}

As illustrated in Fig.\ref{fig:overview}, our framework mainly consists of two components: domain standardization and reference-based photo synthesis. For standardization (details in Section \ref{sec:standarization}), input photos and sketches are converted to standardized edge maps, which bypass the lack of data  issue.
The second part is reference-guided photo synthesis (details in Section \ref{sec:s1}), where synthesized photos are generated based on input sketches and style reference photos.

\subsection{Domain Standardization}
\label{sec:standarization}

\begin{figure}[t!]\centering
\includegraphics[width=\columnwidth]{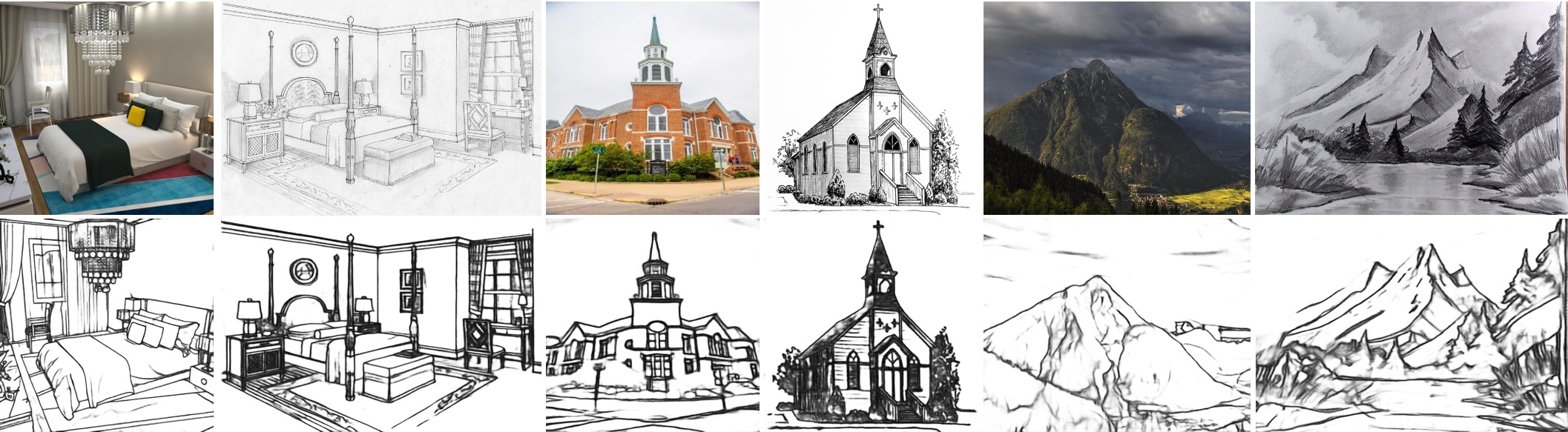}
\caption{The standardization module converts photos and sketches to a standardized domain,  edge maps. After the standardization, edges of photos and sketches share higher similarity, which makes the domain gap between training and evaluation narrower. Within the test set, edges of sketches with different individual styles also a share higher similarity, making the intra-sketch-set discrepancy smaller. }
\vspace{-15pt}
\label{fig:edge}
\end{figure}

Due to the lack of paired sketch-photo datasets, it is intractable for supervised models to synthesize photos from sketches. 
We adopt a similar idea as \cite{wang20203d}, where 
 they converted inputs to a standardized domain, and showed learning from such domain has better performance compared to directly using unprocessed inputs. 

As shown in Fig.\ref{fig:overview}{\bf L}, the standardization can be considered as data prepossessing and is different for training and inference.
During training, we collect a large scale photo dataset of a specific category, e.g., indoor scenes. Each photo is converted to a standardized edge map for later use with an off-the-shelf deep-learning-based edge detector \cite{poma2020dense}. 
During inference, unlike the training, the input is a sketch. We use the same edge detector to convert it to the edge map for later use. Fig.\ref{fig:edge} depicts examples of photo, sketches and their corresponding edges. The standardized edge maps have small domain discrepancies. In addition to narrowing the domain gap between the training and test data, the standardization module during inference could narrow the gap of individual sketching styles (e.g., stroke width), which was also similarly shown in \cite{wang20203d}. Given that edge maps serve as a proxy for real sketches, we slightly abuse the wording of  {\it synthetic sketches} (or omitted as sketches) hereinafter as they may refer to standardized edge maps.

\subsection{Reference-Guided Photo Synthesis}
\label{sec:s1}

Previous works \cite{park2020swapping,viazovetskyi2020stylegan2} show that photos can be encoded to two disentangled representations: content and style representations. 
We extend the concept to sketches and show that they can be encoded to disentangled representations.
Preserving content representation while replacing the sketch style with a real photo style representation could generate a realistic synthesized photo.

The module is trained in two stages. {\bf 1)} Disentangled representation encoding stage learns content and style representations from images via auto-encoding.  {\bf 2)} We further fine-tune the model with sketch-reference-photo triplets, with regularization loss to guarantee the synthesizing quality. 
Our model is inspired by and based on previous arts on disentangled representation learning \cite{park2020swapping} and style transfer \cite{viazovetskyi2020stylegan2}, with novel designs for the goal of scene sketch to photo synthesis.

\begin{figure}[t!]\centering
\includegraphics[width=0.7\columnwidth]{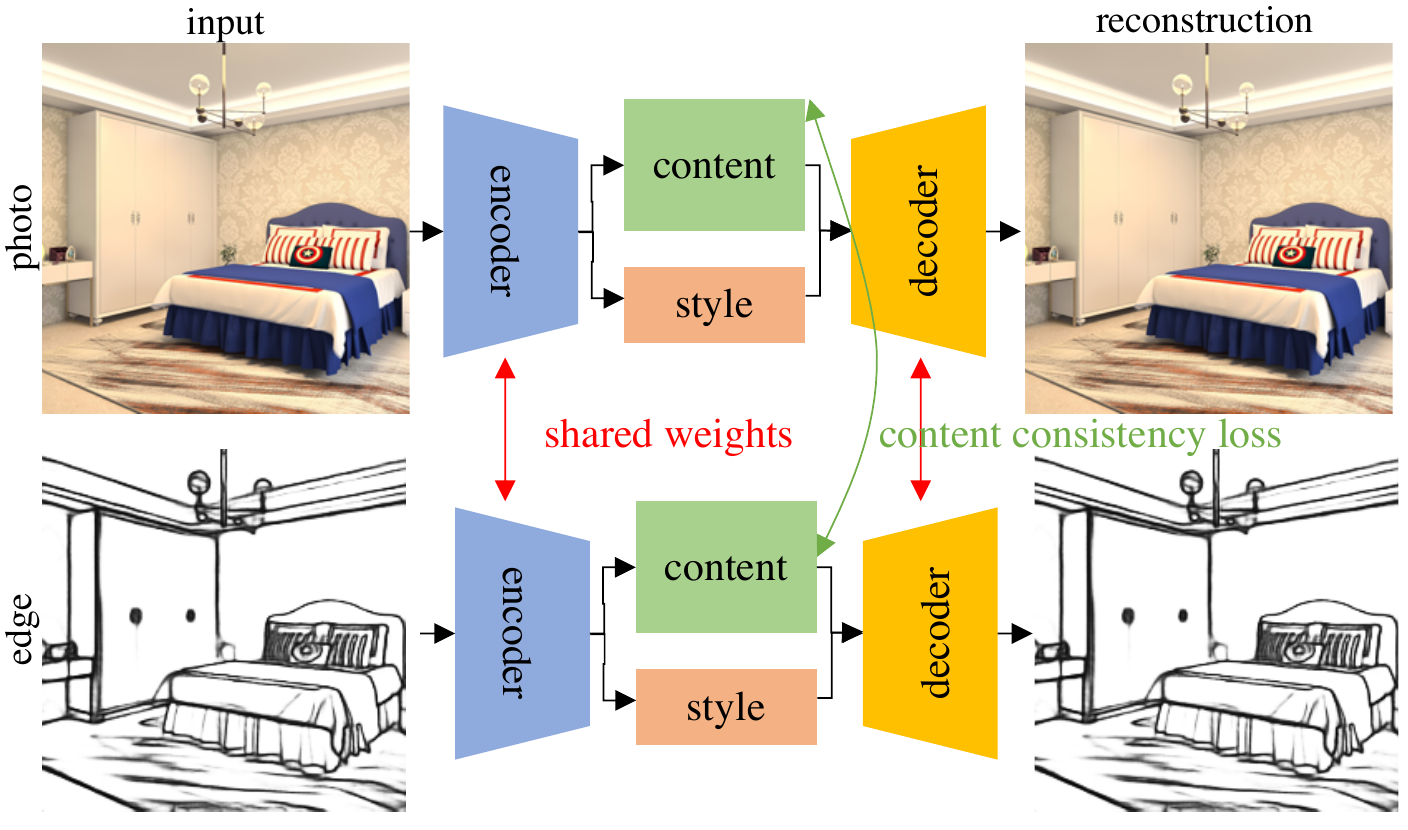}
\vspace{-0.5em}
\caption{ {\it Disentangled representation encoding} is the first stage of the sketch-to-photo synthesis module. 
For each photo, we generate a standardized edge map and form an image pair. Each image of the pair
is encoded as content and style representations by the encoder. 
We add content consistency loss to make content representations of the photo and the edge to be similar.
The representations are then decoded to a reconstructed image by the decoder. The network learns the representations through the auto-encoding process. For the performance of sketch to photo synthesis later, both photos and their corresponding standardized edges are fed to the network for auto-encoding.}
\label{fig:step1}
\end{figure}
\begin{figure}[t!]\centering
\vspace{-1em}
\includegraphics[width=0.8\columnwidth]{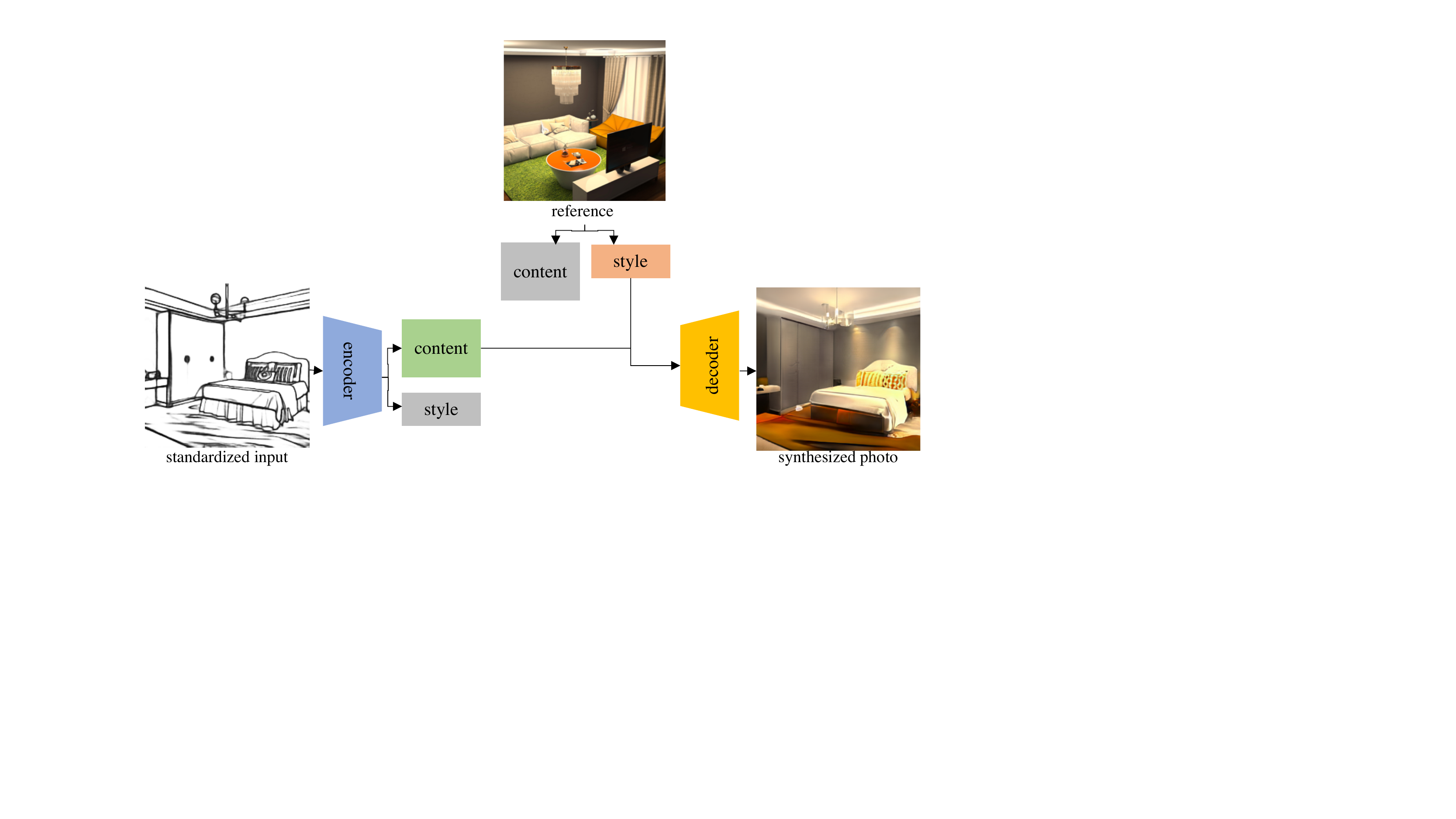}
\caption{ {\it Fine-tuning with sketch-reference-photo triplets} is the second stage of the sketch-to-photo synthesis module. 
The input is a standardized edge map and a reference photo. The model is pre-trained in the  representation encoding phase.
Both the edge map and the reference photo are encoded by the network for content and style representations. The content and representations are fed to the decoder to reconstruct the synthesized photo.
}\vspace{-2em}
\label{fig:step2}
\end{figure}


\noindent  {\bf Disentangled Representation Encoding}.
Fig.\ref{fig:step1} depicts the pipeline of the disentangled representation encoding stage. Denote a pair of input images and its corresponding edge as $\{{\bf x, \bf x^\prime}\}$, 
the encoder as $E$, decoder as $G$, and discriminator as $D$. The encoder encodes input pairs ${\{\bf x, x^\prime\}}$ to two representation pairs, content $\{c_{\bf x}, c_{\bf x^\prime}\}$ and style $\{s_{\bf x}, s_{\bf x^\prime}\}$, i.e., $E({\{\bf x, x^\prime\}}) = \{\{c_{\bf x}, c_{\bf x^\prime}\}, \{s_{\bf x}, s_{\bf x^\prime}\}\}$. From the encoded representations, the decoder reconstructs a photo $G(c_{\bf x}, s_{\bf x})$ and its edge $G(c_{\bf x^\prime}, s_{\bf x^\prime})$. The auto-encoder ensures the reconstructed image pair is similar to the input image pair by the following reconstruction loss in $\ell_1$-norm:
\begin{align}
    \loss_{\text{rec}_1} = \E_{{\bf x} \sim {\bf X}, {\bf x\prime} \sim {\bf X^\prime}}[|{\bf x} - G(c_{\bf x}, s_{\bf x})|+|{\bf x^\prime} - G(c_{\bf x^\prime}, s_{\bf x^\prime})|]
\end{align}
Since the photo and the edge depict the same content, we ask their content representations to be similar in $\ell_1$-norm:
\begin{align}
    \loss_{\text{content}} = \E_{\bf x \sim {\bf X}, {\bf x^\prime }\sim {\bf X^\prime}} [|c_{\bf x} -c_{\bf x^\prime} |]
\end{align}
Further, the adversarial GAN loss \cite{goodfellow2014generative} is required to train discriminator $G$ for realistic reconstructions:
\begin{align}
  \loss_{\text{GAN}_1} = \E_{{\bf x} \sim {\bf X}, {\bf x^\prime} \sim {\bf X^\prime}}[ -\log D(G(c_{\bf x}, s_{\bf x}))-\log D(G(c_{\bf x^\prime}, s_{\bf x^\prime}))]
\end{align}
The final loss is $\loss_{\text{rec}_1}+ \theta \loss_{\text{content}} + \alpha \loss_{\text{GAN}_1}$, where $\theta, \alpha$  are both set to be 0.5.

\noindent {\bf Fine-Tuning with Sketch-Reference-Photo Triplets}. Fig.\ref{fig:step2} depicts the pipeline of the fine-tuning stage. Denote the sketch, reference photo and output synthesized photo as ${\bf x^k, x^r, x^o}$, respectively.  With the pre-trained model from the previous representation learning stage, the encoder is able to encode content and style representations of sketches and photos. The output image is generated by the decoder from the content representation of the sketch $c_{\bf x^k}$, and the style representation of the reference $s_{\bf x^r}$:
\begin{align}
    {\bf x^o} = G(c_{\bf x^k}, s_{\bf x^r})
\end{align}
As the model has been pre-trained in the previous stage for encoding content and style representations, the model has a good starting point for synthesizing photos from sketches. 
To ensure the output image has similar content as the sketch and a similar style as the reference, however, we enforce the following regularization loss on content and style representations in $\ell_1$-norm:
\begin{align}
    \loss_{\text{reg}} = \E_{ {\bf x^k} \sim {\bf X^k}, {\bf x^r} \sim {\bf X^r}, {\bf x^o} \sim {\bf G(c_{\bf X^k}, s_{\bf X^r})} } [|c_{\bf x^o} -c_{\bf x^k} |+|s_{\bf x^o} -s_{\bf x^r}| ]
\end{align}
Additionally, the adversarial GAN loss is required:
\begin{align}
\loss_{\text{GAN}_2} = \E_{{\bf x^k} \sim {\bf X^k}, {\bf x^r} \sim {\bf X^r}}[ -\log D(G(c_{\bf x^k}, s_{\bf x^r}))]
\end{align}
The final loss is $\loss_{\text{reg}} + \beta \loss_{\text{GAN}_2}$, where $\beta$ is set to be 0.5 in the work.

\begin{figure}[t!]\centering
\includegraphics[width=0.95\columnwidth]{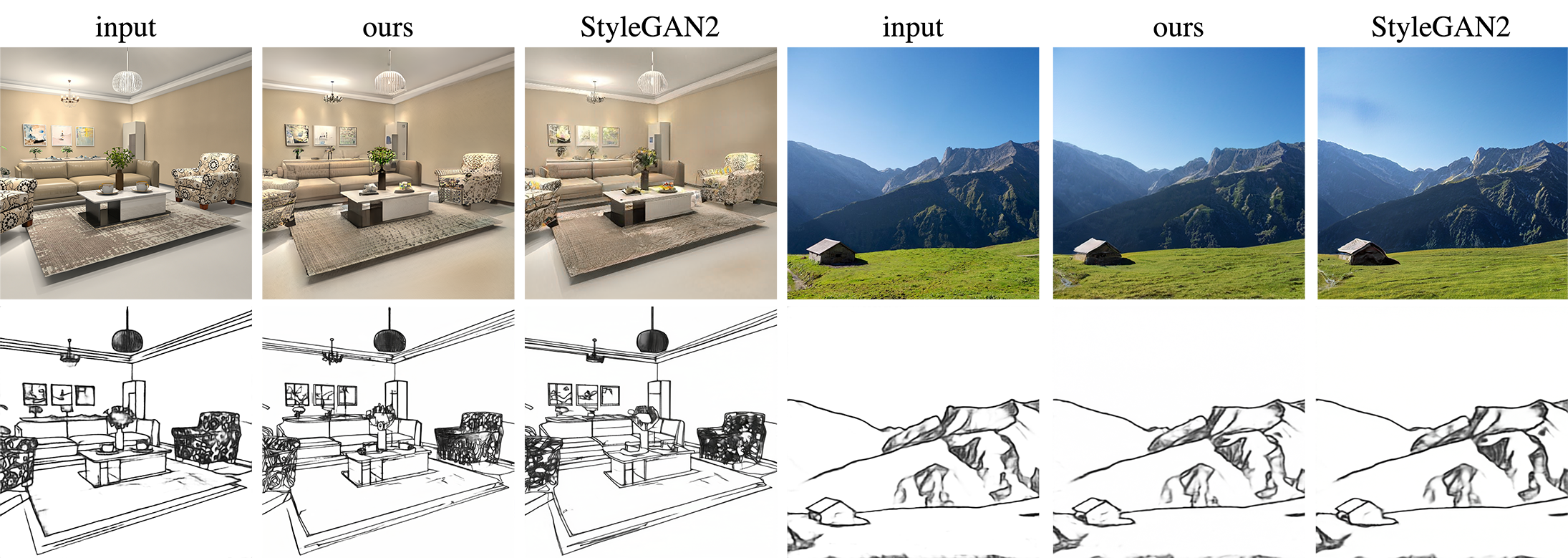}
\vspace{-1em}
\caption{ The reconstruction results of our method and StyleGAN2 \cite{viazovetskyi2020stylegan2}. Images are projected into embedding spaces for ours and StyleGAN2\cite{viazovetskyi2020stylegan2}. Both photos and standardized edges are fed to the network for reconstruction. The high faithfulness in reconstruction demonstrates that the learned content and style representations are effective.
}
\vspace{-1em}
\label{fig:recon}
\end{figure}

\begin{figure*}[htb]\centering
\vspace{-1em}
\includegraphics[width=\textwidth]{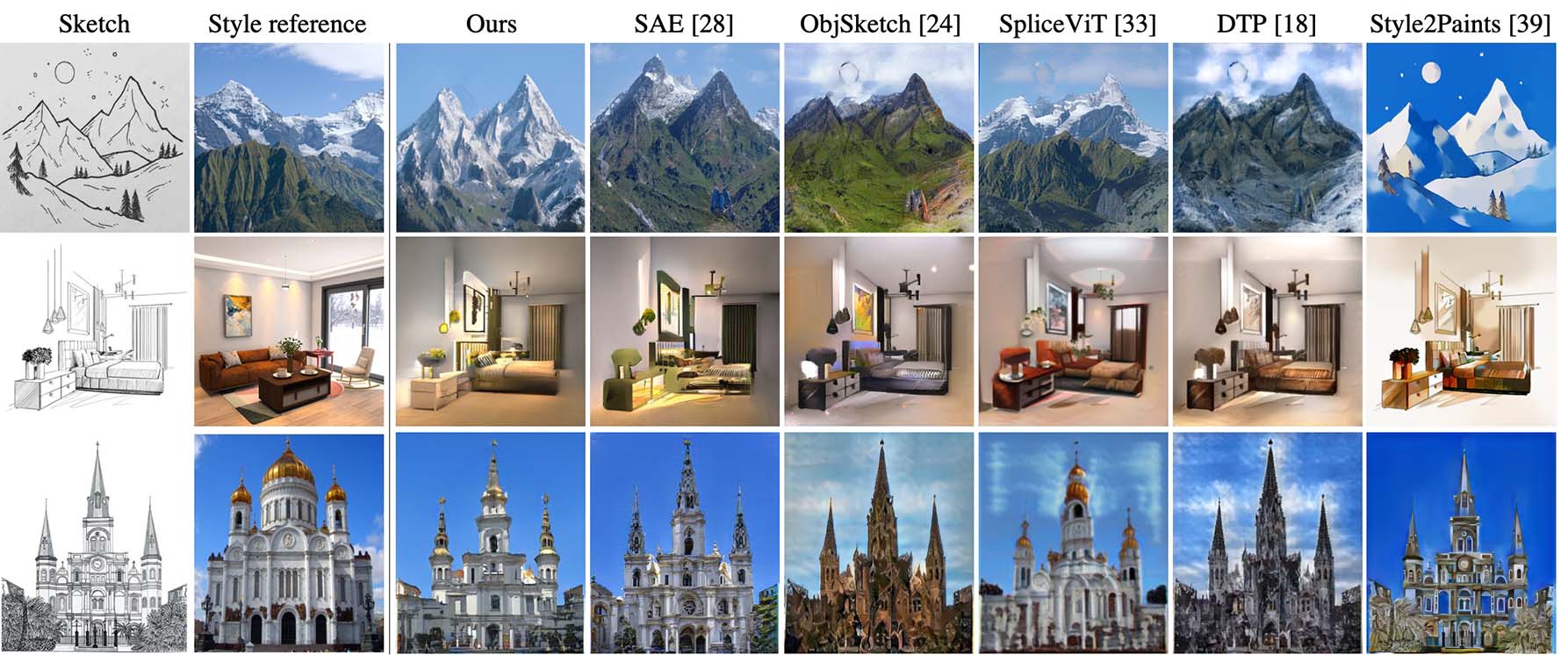}
\caption{Various baseline photo syntheses from sketches with style guidance. 
Note that SpliceViT \cite{tumanyan2022splicing} and DTP \cite{kim2021deep} are designed for test-time optimization and are not trained on the full dataset, making them disadvantageous to other methods. All other methods are trained on the same dataset with a similar iteration as the proposed method. Style2Paints is designed to synthesize painting, not realistic photos.
Our model synthesizes photos that share a similar content as the sketch and a similar visual style as the style photo reference.
}
\vspace{-1em}
\label{fig:baseline}
\end{figure*}

\begin{table*}[t]
    \caption{ \small {\bf (a)} Reconstruction performance measured in LPIPS ($\downarrow$) \cite{zhang2018unreasonable}. Images are projected into embedding spaces for ours and StyleGAN2\cite{viazovetskyi2020stylegan2}. We reconstruct photos and edges with a similar performance as StyleGAN2\cite{viazovetskyi2020stylegan2}, demonstrating the disentanglement to content and style representations is effective. {\bf (b)} Reference-guided sketch to photo synthesis performance measured in FID ($\downarrow$) \cite{heusel2017gans}. Our method outperforms other baseline methods in all three categories.}
        \vspace{-1em}

\label{tab:general}
    \begin{subtable}{.46\linewidth}
      \centering
        \caption{}
\resizebox{\columnwidth}{!}{\begin{tabular}{c|c|c|c|c|c}
\hline
\rowcolor{LightCyan}  input                  & method    & indoor         & church         & mountain       & mean           \\ \hline
\multirow{2}{*}{photo} & ours      & \textbf{0.254} & \textbf{0.214} & \textbf{0.221} & \textbf{0.229} \\ \cline{2-6} 
                       & StyleGAN2 & 0.256          & 0.220          & 0.224          & 0.233          \\ \hline
\multirow{2}{*}{edge}  & ours      & \textbf{0.180} & \textbf{0.166} & \textbf{0.171} & \textbf{0.172} \\ \cline{2-6} 
                       & StyleGAN2 & 0.161          & 0.188          & 0.173          & 0.174          \\ \hline
\end{tabular}}
    \end{subtable}%
    \hfill
    \begin{subtable}{.46\linewidth}
      \centering
        \caption{}
\resizebox{\columnwidth}{!}{\begin{tabular}{>{\columncolor[gray]{0.95}}c|c|c|c|c}
\hline
\rowcolor{LightCyan}  FID ($\downarrow$)   & indoor & church & mountain & mean          \\ \hline
ours         & \textbf{105.5}                    & \textbf{48.7}               & \textbf{73.8}                 & \textbf{76.0} \\ \hline
SAE~\cite{park2020swapping}         & 107.7                             & 52.4                        & 74.1                          & 78.1          \\ \hline
ObjSketch~\cite{liu2020unsupervised}   & 136.5                             & 62.1                        & 95.4                          & 98.0          \\ \hline
SpliceViT~\cite{tumanyan2022splicing}       & 204.2                             & 119.7                       & 140.7                         & 154.9         \\ \hline
DTP~\cite{kim2021deep}          & 205.2                             & 124.2                       & 143.5                         & 157.6         \\ \hline
Style2Paints~\cite{Filling2021zhang} & 254.2                             & 217.3                       & 247.7                         & 239.7         \\ \hline
\end{tabular}}
    \end{subtable} 
    \vspace{-1em}
\end{table*}

\section{Experimental Results}

\subsection{Network Architectures and Training Details}
{\bf Network Architectures}. Images are fed to the encoder to obtain content and style representations. 
First, images go through 4 down-sampling residual blocks \cite{he2016deep} to obtain an intermediate representation. The intermediate representation is fed to another convolution layer to obtain the content representation with a spatial size of $16 \times 16$. The intermediate representation is also fed to another two convolution layers to obtain a style representation/vector dimension of $2048$. The decoder consists of 4 up-sampling residual blocks. The style representation is injected to the decoder convolution layers with weight modulation techniques described in StyleGAN2 \cite{viazovetskyi2020stylegan2}. The discriminator is the same as that of StyleGAN2. 

\noindent {\bf Hyper-Parameters and Training Schedules}. For representation encoding, the initial learning rate is 2e-3. We use Adam optimizer \cite{kingma2015adam} with $\beta=(0, 0.99)$. For fine-tuning, we start from the previously pre-trained model. The training schedule stays the same with the initial learning rate being 4e-4. The entire training time for the 3D-front indoor scene dataset is 7 days on 4 V100 GPUs.

\noindent {\bf Baselines}. We follow the released code and the same settings of all baseline methods and retrain on datasets used in the paper. Specifically, some baselines \cite{park2020swapping,liu2020unsupervised,tumanyan2022splicing,kim2021deep} only work on photos, but not sketches. We use a gray-scale images as a proxy to ensure the photo synthesis quality. Specifically, we first train a sketch to gray-scale photo model using the same setting as step 1 of \cite{liu2020unsupervised}, where the input to the model is a standardized sketch. The generated gray-scale photo is then used to train a gray-scale to color photo model with the same setting of the baseline methods. SpliceViT \cite{tumanyan2022splicing} and DTP \cite{kim2021deep} are designed for test-time optimization and are not trained on the entire dataset. All other baseline methods are trained on the same dataset as the proposed method with a similar iteration.

\vspace{-1em}
\subsection{Datasets}

We train on the following scene photo datasets:
 \textbf{1) 3D-Front Indoor Scene} \cite{fu20213d} consists of 14,761 training and 5,479 validation photos. They are rendered with Blender from synthetic indoor scenes including bedrooms and living rooms. Photos are resized to 286 and randomly cropped to 256 during training.
\textbf{2) LSUN Church} \cite{yu2015lsun} consists of 126,227 photos of outdoor churches. We randomly sample 25,255 photos as the validation set. Photos are resized to 286 and randomly cropped to 256 during training.
 \textbf{3) GeoPose3K Mountain Landscape} \cite{brejcha2017geopose3k} has 3,114 mountain landscape photos. 623 photos are randomly sampled for validation. Training photos are resized to 572 and randomly cropped.

For evaluation, we collect a \textbf{Scene Sketch Evaluation Set}. For each category (indoor scenes, mountain and church), we collect 50 sketches from the Internet, respectively. The sketches are collected with an intention to cover various sketching styles, e.g. different levels of line width, geometric distortion, use of shading, etc.

\subsection{Representation Encoding} 

With effective learned representation, the model could reconstruct photos or sketches with high quality. We evaluate reconstruction performance in LPIPS \cite{zhang2018unreasonable}.

Table \ref{tab:general}{\bf a} reports the LPIPS distance of reconstructed and input photos and synthetic sketches of our stage 1 model and StyleGAN2 \cite{viazovetskyi2020stylegan2}. Fig.\ref{fig:recon} depicts several examples of the input and reconstruction. Our representation encoding model has a slightly better reconstruction performance compared to StyleGAN2, indicating the learned content and style representations are adequate and ready for further fine-tuning with sketch-reference-photo pairs.

\begin{figure*}[t!]\centering
\includegraphics[width=\textwidth]{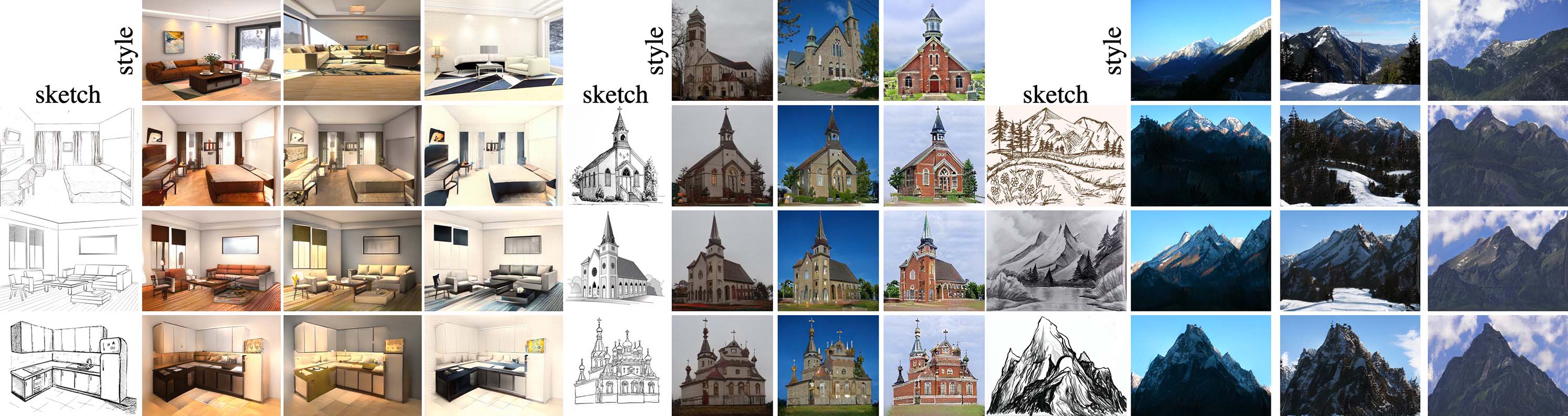}
\vspace{-1em}
\caption{ The indoor scene, church and mountain sketch to photo synthesis with different references. We synthesize high-fidelity scene photos with similar content as the sketch and similar style as the reference photos.
}
\label{fig:style1}
\end{figure*}

\subsection{Photo Synthesis }

We evaluate the photo synthesis performance of our method and baselines in terms of photo-realism. We calculate the Fréchet inception distance (FID) \cite{heusel2017gans} between the synthesized photo set and the training photo set for each category (Table \ref{tab:general}{\bf b}). Our method outperforms other baselines under the FID metric. Fig.\ref{fig:baseline} depicts synthesis results of our method and baselines. Note that SpliceViT \cite{tumanyan2022splicing} and DTP \cite{kim2021deep} designed for test-time optimization and was not trained on the full dataset, making it disadvantageous to other methods. Style2Paints is designed to synthesizing painting, not realistic photos. We however include it as it is one of the few works that study synthesizing from scene sketches. Our synthesis result outperforms all other methods, with SAE \cite{park2020swapping} being the second.
As for if the content of the output photo matches with the input sketch or if the style matches with the reference photo, we provide human perceptual evaluation in Section \ref{sec:us}.

We also provide more visualization of our synthesis results of indoor scenes, churches and mountains in Fig.\ref{fig:style1}.

\subsection{Human Perceptual Study}
\label{sec:us}


\begin{table*}[t]
    \caption{ \small A human perceptual study of the synthesized photos. 
    {\bf (a)} The fooling rate of our synthesized model over real photos measures the realism of the generation.
    {\bf (b)} User preference on which method synthesizes photos that depicts more similar content to the sketch.
    {\bf (c)} User preference on which method synthesizes photos that depicts more similar visual style to the reference photo.
Compared with \cite{park2020swapping}, we have a higher fooling rate over real photos, better content and style matching preference rate.}
        \vspace{-1em}

\label{tab:user}
    \begin{subtable}{.32\linewidth}
      \centering
        \caption{Fooling rate ($\uparrow$) }
\resizebox{\columnwidth}{!}{\begin{tabular}{>{\columncolor[gray]{0.95}}c|c|c|c|c}
\hline
\rowcolor{LightCyan} (\%) & indoor scene & church  & mountain & mean    \\ \hline
ours & {\bf 25.00}      & {\bf 44.3} & {\bf 48.9}  & {\bf 39.4} \\ \hline
SAE~\cite{park2020swapping} & 10.0       & 6.6  & 20.0  & 12.2 \\ \hline
\end{tabular}}
    \end{subtable}%
    \hfill
    \begin{subtable}{.32\linewidth}
      \centering
        \caption{Content matching ($\uparrow$)}
\resizebox{\columnwidth}{!}{\begin{tabular}{>{\columncolor[gray]{0.95}}c|c|c|c|c}
\hline
\rowcolor{LightCyan} (\%) & indoor scene    & church          & mountain        & mean             \\ \hline
ours & \textbf{80.1} & \textbf{92.1} & \textbf{75.0} & \textbf{82.4} \\ \hline
SAE~\cite{park2020swapping}  & 19.9         & 7.9           & 25.0         & 17.6          \\ \hline
\end{tabular}}
    \end{subtable} 
    \hfill
    \begin{subtable}{.32\linewidth}
      \centering
        \caption{Style matching ($\uparrow$)}
\resizebox{\columnwidth}{!}{\begin{tabular}{>{\columncolor[gray]{0.95}}c|c|c|c|c}
\hline
\rowcolor{LightCyan} (\%) & indoor scene  & church        & mountain      & mean           \\ \hline
ours & \textbf{61.9} & \textbf{90.9} & \textbf{71.0} & \textbf{74.6} \\ \hline
SAE~\cite{park2020swapping}  & 38.1          & 9.1          & 29.0          & 25.4         \\ \hline
\end{tabular}}
    \end{subtable}
    \vspace{-1em}
\end{table*}

We conduct a human perceptual study to evaluate the realism of synthesized photos, and if synthesized photos match contents and  styles as desired. We only evaluate our method and SAE \cite{park2020swapping}, the second best-performing synthesis method, due to limited resources.

We create a survey consisting of three parts: photorealism, content matching with sketches and style matching with reference photos. As guidance to the participants, we state our research purpose at the beginning of the survey. For each part, a detailed description and an example question with answers and explanations are provided for the participant's reference.
The order of our results, baseline results, and real images are randomly shuffled in the survey to minimize the potential bias from the participant.
Each part consists of 13 questions, with one question being a {\it bait question} with an obvious answer. The bait question is designed to check if the participant is paying attention and if the answers are reliable. There are in total 51 participants, with 1 being ruled out due to failing one of the bait questions. Thus we finally collect 1,950 valid human judgments.

To evaluate the photorealism, we randomly select synthesized photos of ours and SAE evenly from three categories.
Both methods use the same input sketch and reference photo. For each synthesized photo, we use Google's search by image feature to find the most similar real photo and ask participants which one they think looks more like a real photo.
We then calculate the percentage of participants being fooled. Note that the fooling rate of random guessing is 50\%.
Table \ref{tab:user}{\bf a} reports the fooling rate of our method and SAE. Ours is 27\% higher than SAE. Specifically, for churches and mountains, ours achieves a fooling rate over 44\%: the generated photos are almost indistinguishable from real photos.

To evaluate if the synthesized photos match the content of the input sketch, we show participants an input sketch and two synthesized results from our method and SAE, and ask them to pick one that has the most similar content as the sketch. Table \ref{tab:user}{\bf b} reports the preference rate of ours over SAE. We achieve 82\% on average preference rate, well outperforming the baseline.

To evaluate if the synthesized photos match the style of the reference photo, we show participants a reference photo and two synthesized results from our method and SAE, and ask them to pick one that has the most similar style to the sketch. Table \ref{tab:user}{\bf c} reports the preference rate of ours over SAE. We achieve a 75\% average preference rate, well outperforming the baseline.

\begin{figure}[t!]\centering
\includegraphics[width=0.6\columnwidth]{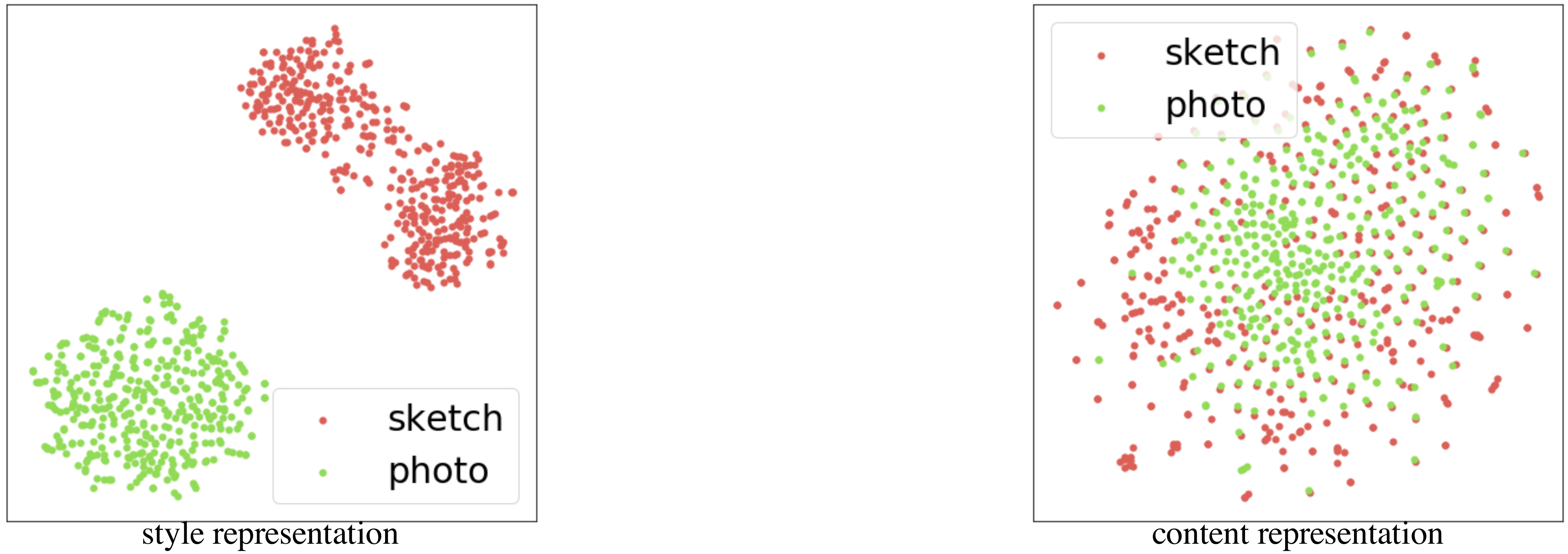}
\vspace{-0.5em}
\caption{ The style representations of sketches and photos are well separated, while the content representations of sketches and photos are tangled together. We visualize learned content and style representations of sketches and photos  with T-SNE \cite{van2008visualizing}. The results show that sketches and photos share the content space and it is appropriate to train on photos and transfer knowledge to sketches.
}
\label{fig:style_vis}
\end{figure}
\begin{figure*}[t!]\centering
\includegraphics[width=\textwidth]{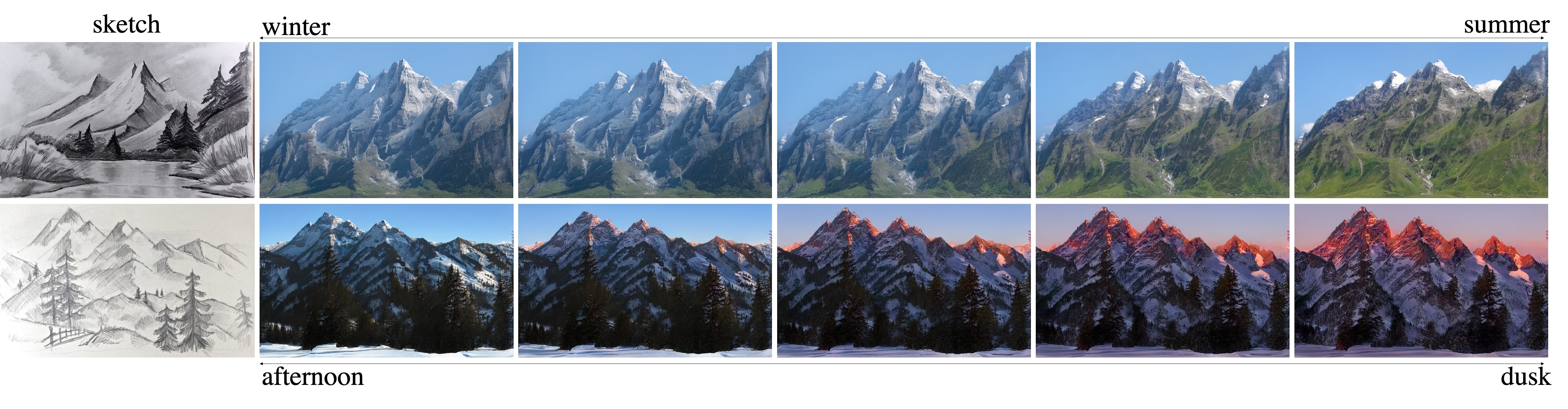}
\vspace{-2em}
\caption{ Sketch to photo synthesis with combined style representations of two references. 
We encode style representations from two photos, e.g. a winter photo and a summer photo. By increasing the weight of the summer image and decreasing that of the winter image, the synthesized photo from the sketch gradually changes from winter appearance to summer appearance.
}
\label{fig:interp}
\end{figure*}

\subsection{Photo Editing Through Sketch}
\label{sec:edit}
\begin{figure}[t!]\centering
\includegraphics[width=\columnwidth]{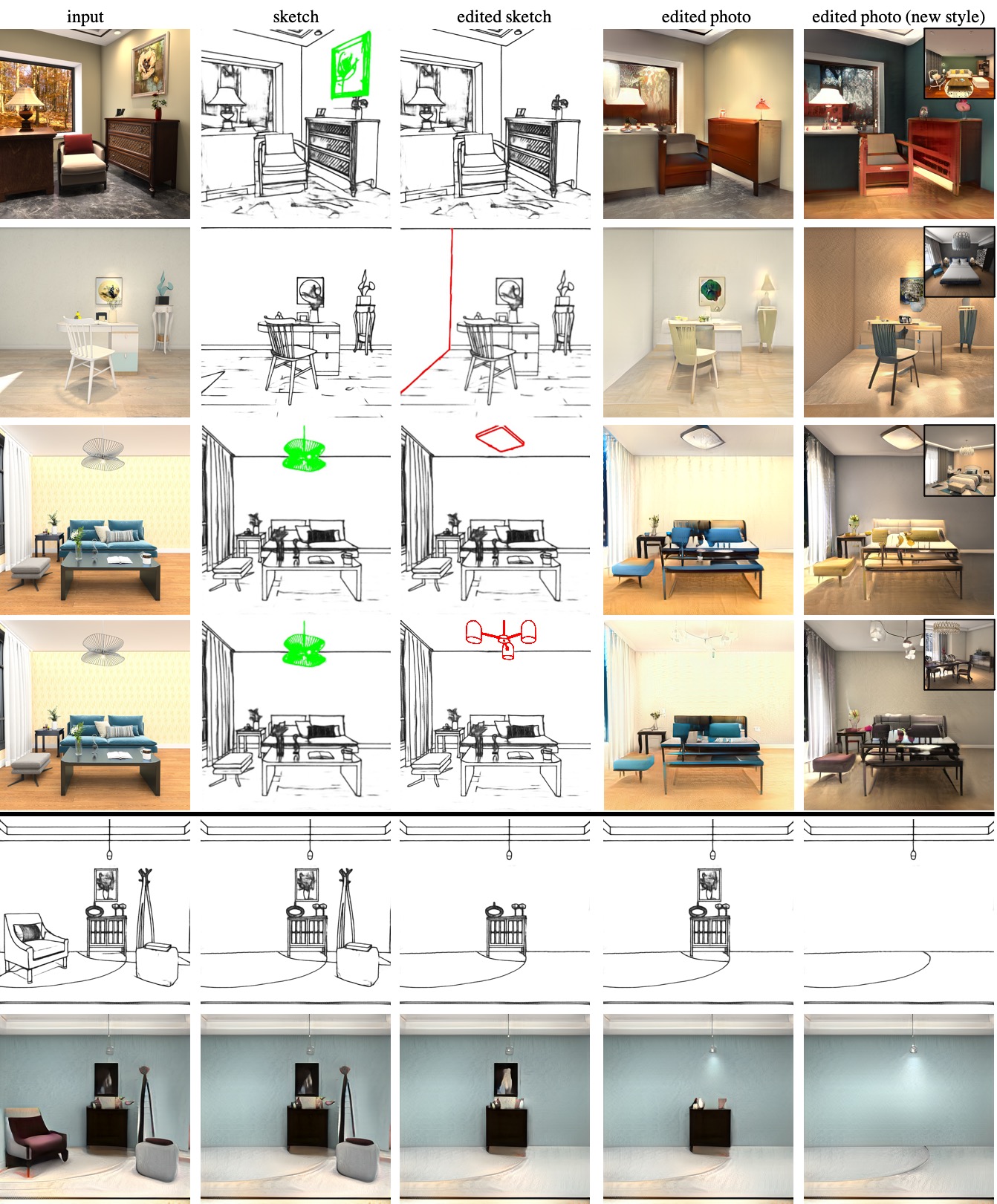}
\vspace{-1.5em}
\caption{ Photo editing and style transfer via sketches. {\it Upper:} Given an input image, we first convert it to a standardized edge map. We then \textcolor{red}{add} or \textcolor{green}{remove} strokes in the edge map and convert it back to a photo. The visual style of the photo could also be changed with a reference photo (top right). {\it Lower:} Sequential editing by gradually removing strokes.}
\label{fig:edit}
\vspace{-1.2em}
\end{figure}

As depicted in Fig.\ref{fig:edit}, given an input photo, we convert it to a standardized edge map (where we refer as sketch for simplicity). Users could add and remove strokes to edit the photo. We also show the possibility of sequential editing in the figure. We evaluate the photo editing performance for the indoor scene validation dataset, and the FID \cite{heusel2017gans} of edited images to the training set is 69.2. One limitation is that the content in the unmodified region of a given photo may not be well preserved as the edited photo is solely generated from the edge map.

\subsection{Analysis and Ablation Studies}

\textbf{Analysis of Style Representations}. We visualize the learned content and style representations of photos and sketches using T-SNE \cite{van2008visualizing} in Fig.\ref{fig:style_vis}: style representations of sketches and photos are well separated, while content representations of sketches and photos are not separable. This verifies the grounding of the method: the content representations of sketches and photos can be shared, while the style representations for the two are different. Thus, combining the content representation of a sketch and style representation of a photo could decode a realistic synthesized photo.

\noindent \textbf{Style Interpolation}.We study if the reference style can be a combination of style of two different reference images ${\bf x^{r_1}}$ and ${\bf x^{r_2}}$. Suppose their style representations are ${\bf s_{x^{r_1}}}$ and ${\bf s_{x^{r_2}}}$. The combined representation $s_{\text{combined}} = \gamma {\bf s_{x^{r_1}}} + (1-\gamma) {\bf s_{x^{r_2}}}$, where $\gamma \in [0, 1]$. By adjusting $\gamma$, we synthesize photos with a combined style from both reference images. 
Fig.\ref{fig:interp} depicts examples of mountain sketch to photo synthesis with combined styles from two different reference images. By adjusting $\gamma$, the synthesized photos have a continuous interpolation from winter to summer, and afternoon to dusk.

\begin{table}[t!]
\begin{center}
\centering
\vspace{-1em}
\caption{ Ablation studies on the fine-tuning stage, content and style regularization loss for indoor scenes in FID ($
\downarrow$) \cite{heusel2017gans} distance. Having both stage 2 fine-tuning and the regularization loss gives the best result.}
\label{tab:ablation}
\resizebox{0.9\columnwidth}{!}{
\begin{tabular}{c|c|c|c}
\hline
\rowcolor{LightCyan}  no fine-tune                   & fine-tune+style loss     & fine-tune+content loss     & fine-tune+all loss         \\ \hline
 107.9 & 107.0 & 106.1 & {\bf 105.5} \\ \hline
\end{tabular}

}
\vspace{-2em}
\end{center}
\end{table}

\noindent \textbf{Fine-Tuning Model}. One of the novelty is that we propose the fine-tuning with sketch-reference-photo triplets for the task. 
We evaluate if the fine-tuning is necessary by removing the fine-tuning stage.
As reported in Table \ref{tab:ablation}, removing the model fine-tuning leads to 2.4 worse results in the FID metric.

\noindent \textbf{Content and Style Regularization Loss}. 
We study if the regularization loss at the fine-tuning stage is effective. We study the function of the content loss ($|c_{\bf x^o} -c_{\bf x^k} |$) and style loss ($|s_{\bf x^o} -s_{\bf x^r}|$) respectively. As reported in Table \ref{tab:ablation}, removing the content regularization loss leads to 1.5 worse results in FID metric, and removing the style loss leads to 0.6 worse results. This verifies the effectiveness of the proposed regularization loss.
\vspace{-1em}
\section{Summary}
We propose a reference-guided framework for photo synthesis from scene sketches.
We first convert all input photos and sketches to standardized edge maps, allowing the model to learn in unsupervised setting without the need of real sketches or sketch-photo pairs. Sequentially, the standardized input and reference image are disentangled into content and style components to synthesize new hybrid image that preserves the content of standardized input while transferring the style of reference image.
Extensive experiments demonstrate that our method can generate and edit a realistic photo from a user's scene sketch with a reference photo as style guidance, surpassing the previous approaches on three benchmarks.

A major insight of this work is that, we learn to synthesize scene structures directly from the vast amount of readily-available photos, rather than synthesizing and combining individual objects. Rather than worrying about the acclimated errors from sketch-based object detection, photo synthesis and spatial combination for the final output, we treat the scene sketches as a whole and learn the holistic structures for photo synthesis.

One limitation is that the deep-learning based standardization step could eliminate strokes that reflect the details of the scene, or misinterpret the strokes as textures. Future work could study a sketch-to-edge standardization process that preserves higher fidelity of the sketch. Another limitation lies in the sketch-based photo editing - the unchanged regions of a given photo may not be well preserved. This is due to the model takes sketch as the only input. Future work could improve the performance by taking the original photo into consideration. 

{\noindent \small {\bf Acknowledgements}: This research was supported, in part, by BAIR-Amazon Commons and AWS. We thank Yubei Chen for helpful discussions. We thank Tian Qin for providing some scene sketches used in the study. We thank Li Tang, Lu Yuan, Martin Zhai, Xingchen Liu, Karl Hillesland, Amin Kheradmand, Nasim Souly, Charlotte Wang, Valerie Moss and other anonymous participants in our human perceptual study.}

\clearpage
\bibliographystyle{splncs04}
\bibliography{sample-base}

\end{document}